%% file: main.tex
\documentclass[letterpaper, 10 pt, conference]{IEEEconf}

\IEEEoverridecommandlockouts                              

\usepackage{graphics} 
\usepackage{mathtools}
\usepackage{times} 
\usepackage{amsmath} 
\usepackage{amssymb}  
\usepackage{xcolor}
\usepackage[utf8]{inputenc}
\usepackage{comment}
\usepackage{algorithm,algorithmic}
\usepackage{cite}
\newcommand{\R}{\mathbb{R}}
\newcommand{\Poin}{\mathcal{P}}
\newcommand{\x}{\mathbf{x}}
\newcommand{\e}{\mathbf{e}}

\newcommand{\km}{{k-1}}
\newcommand{\kp}{{k+1}}
\newcommand{\hTheta}{\hat{\Theta}}

\newcommand{\datafigurewidth}{\linewidth}

\usepackage{CJKutf8}

\DeclareMathAlphabet{\mathcal}{OMS}{cmsy}{m}{n}

\title{\LARGE \bf
Data-driven Adaptation for Robust Bipedal Locomotion \\with Step-to-Step Dynamics}

\author{Min Dai$^1$, Xiaobin Xiong$^2$, Jaemin Lee$^1$, and Aaron D. Ames$^1$
\thanks{ $^1$ Department of Mechanical and Civil Engineering, California Institute of Technology, Pasadena, CA 91125 USA.
        {\tt\small \{mdai, jaemin87, ames\}@caltech.edu}.  }%
\thanks{ $^2$ Department of Mechanical Engineering, University of Wisconsin-Madison, Madison, WI 53706 USA.
        {\tt\small xiaobin.xiong@wisc.edu}.  }%
\thanks{ This work is supported by NSF NRI award 1924526 and NSF CMMI award 1923239.  }%
}
\begin{document}

\maketitle
\thispagestyle{empty}
\pagestyle{empty}

\input{intro}

\input{prelim}

\input{methodsnew}

\input{results}



\bibliographystyle{IEEEtran}

\bibliography{references,misc}

\end{document}

%% file: intro.tex
\begin{abstract}
This paper presents an \emph{online} framework for synthesizing agile locomotion for bipedal robots that adapts to unknown environments, modeling errors, and external disturbances. 
To this end, we leverage step-to-step (S2S) dynamics which has proven effective in realizing dynamic walking on underactuated robots---assuming known dynamics and environments.
This paper considers the case of uncertain models and environments and presents a data-driven representation of the S2S dynamics that can be learned via an adaptive control approach that is both data-efficient and easy to implement. The learned S2S controller generates desired discrete foot placement, which is then realized on the full-order dynamics of the bipedal robot by tracking desired outputs synthesized from the given foot placement. The benefits of the proposed approach are twofold.  
First, it improves the ability of the robot to walk at a given desired velocity when compared to the non-adaptive baseline controller.  
Second, the data-driven approach enables stable and agile locomotion under the effect of various unknown disturbances: additional unmodeled payload, large robot model errors, external disturbance forces, biased velocity estimation, and sloped terrains.  
This is demonstrated through in-depth evaluation with a high-fidelity simulation of the bipedal robot Cassie subject to the aforementioned disturbances \cite{video-supp}. 
\end{abstract}

\section{Introduction}

Over the decades, a wide range of control methods has been proposed to realize stable bipedal walking. These methods can be roughly divided into two categories: model-based and data-driven. From the model-based perspective, popular methods include zero moment point \cite{vukobratovic_zero-moment_2004}, hybrid zero dynamics \cite{westervelt_hybrid_2003,grizzle_models_2014}, and various reduced-order model based approaches \cite{rezazadeh_control_2020, pratt_capturability-based_2012, gong_angular_2021,garcia_mpc-based_2021}. The linear inverted pendulum (LIP) model \cite{kajita_3d_2001} and its variants are one of the most widely used reduced-order models. Its linearity allows for fast execution of online planning techniques such as model predictive controllers \cite{garcia_mpc-based_2021, kunal2022MPC, khadiv_walking_2020}. 
However, these reduced-order models are often times ``generic'' models, i.e. they are not specific to any particular robots or environment, which can limit the dynamic behaviors these approaches can realize.

More recently, the advancement in computation power opens up the possibility for pure data-based approaches. Model-free methods such as reinforcement learning (RL) have shown some success in real-world dynamic robotic locomotion \cite{siekmann_blind_2021, li_reinforcement_2021,dao_sim--real_2022}. The ability to easily randomize parameters of dynamics and environment in a fast simulation environment allows RL algorithms to successfully find policies that are robust to different types of loads and different terrains and mitigate the problem of sim-to-real transfer to some extent. However, the amount of data required can be massive, and the black box algorithm is very sensitive to reward design, which is still an art through trial and error.

\begin{figure}[t]
    \centering
    \includegraphics[width = 1 \linewidth]{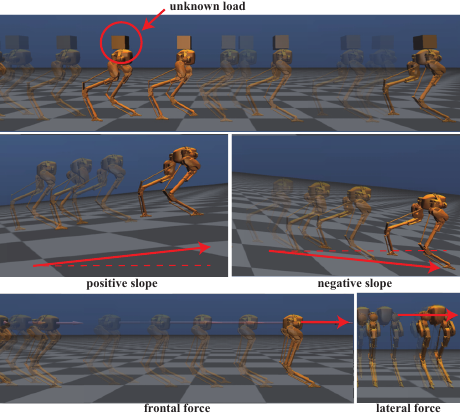}
    \vspace{-6mm}
    \caption{Robot Cassie walking with unknown load, on tilted floors, and  with drag forces using the proposed adaptive S2S based control.}
    \vspace{-4mm}
    \label{fig::overview}
\end{figure}


In attempts to improve robotic walking performance and robustness in a data-efficient way, combinations of model-based and data-driven approaches have been studied. For example, \cite{csomay-shanklin_learning_2022, yang_bayesian_2022, castillo_velocity_2020} uses data to optimize existing controller parameters. Data-driven reduced-order models are considered in \cite{chen_optimal_2020, hernandez-hinojosa_optimal_2021}, where a reduced-order model of the hybrid dynamics or the step-to-step (S2S) dynamics for a specific robot is obtained from simulation data. Using a similar concept, a linear discrete reduced-order S2S dynamics model is learned in \cite{xiong_robust_2021,paredes2022resolved} by extending a Hybrid-LIP  based S2S approximation in \cite{xiong_3-d_2022}: both of which have a particular input-state dynamics structure, where they consider the step size as the input to control horizontal center of mass (COM) states of bipedal walking. Additionally, the subspace approach in \cite{hou_model-based_2013} has been used to obtain a data-driven reduced-order model from the experimental data using the Hankel matrix \cite{fawcett_toward_2022}, which ultimately realized quadrupedal locomotion experimentally. Although these methods use a robot- or problem-specific reduced-order model that can be used efficiently for online planning, they require offline data collection and computation to generate the data-driven approximation. 


This paper presents a novel approach to enhancing the robustness of bipedal locomotion in real-time using an online data-driven approach. The proposed method leverages classical adaptive control methodologies \cite{ioannou_adaptive_2006,swarnkar_adaptive_2014, zhang_review_2017}, known for their ability to operate online and efficiently utilize data. In addition, S2S dynamics, a key component of the approach, is employed to control underactuated walking robots. The successful testing of this method on the Cassie biped \cite{xiong_3-d_2022} and in offline data-driven approaches \cite{xiong_robust_2021} demonstrates its effectiveness. The proposed approach employs indirect adaptive control to learn the S2S dynamic model of the robot and then synthesizes controllers to stabilize the learned S2S model. The online process generates outputs that are tracked by the low-level controller to achieve robust locomotion.

The main contributions of this paper include the following. Firstly, our proposed approach enhances the velocity tracking performance of a bipedal robot while implementing dynamic walking. We substantiate this improvement by showcasing diverse conditions, including unknown payloads, inaccurate system models, external perturbations, biased velocity estimation, and sloped terrain, as depicted in Fig. \ref{fig::overview}. Secondly, our method enables the robot to resist larger disturbances during dynamic stepping, compared to a baseline approach using H-LIP. Lastly, our approach is more responsive than offline data-driven approaches, as we update the data online.
These conclusions are verified with in-depth testing on a high-fidelity simulation of the Cassie bipedal robot\cite{cassie}.

The rest of the paper is organized as follows. Section \ref{sec::S2S} introduces the notion of S2S dynamics of walking and the previous reduced-order model based framework for walking synthesis. Section \ref{sec::adaptive} summarizes the adaptive control methods and the formulations we used for bipedal walking synthesis. We then present the results and evaluate the performance under different circumstances in section \ref{sec::results}. Finally, the conclusion is given in section \ref{sec::conclusion}.

%% file: prelim.tex
\section{S2S dynamics Approximation based \\
Bipedal Walking}\label{sec::S2S}
\subsection{Robot Hybrid Dynamics and Step-to-step Dynamics}
Bipedal robotic walking is typically modeled as a hybrid dynamical system \cite{grizzle_models_2014} that undergoes continuous dynamics and discrete transitions: the continuous dynamics are derived from Lagrangian mechanics, and the discrete foot-ground impacts are assumed plastic and instantaneous. The hybrid control system is described by:
\begin{align}
    \mathcal{HC} :\begin{cases} \dot{x} = f_n(x,\tau),\\
    x^+ = \Delta_{n \rightarrow n+1}(x^-),\end{cases}
\end{align}
where $x$ is the state variable, $f_n$ denotes the continuous nonlinear dynamics in domain $n$, $\tau$ is the actuation torque, $\Delta_{n \rightarrow n+1}$ represents the discrete transition from the domain $n$ to the domain $n+1$, and the superscripts $^{+/-}$ indicate the state after/before the discrete transitions, respectively.

The bipedal walking motion naturally produces step-level discrete dynamics, i.e., the step-to-step (S2S) dynamics \cite{bhounsule_low-bandwidth_2014}. In control language, the S2S dynamics is the Poincar\'e return map of the hybrid dynamics at the Poincar\'e section, which is typically selected at the surface of the impact event. Let $x^-_k$ be the pre-impact state of the $k$-th foot-strike with the ground. Assuming the existence of the next foot-strike event, the S2S dynamics of the robot can be written as $x^-_{k+1} = \mathcal{P}(x_k^-,\tau(t))$, 
where $\tau(t)$ is the continuous joint actuation torque with $t$ represents the continuous time. For walking, we primarily care the stability of the weakly-actuated horizontal COM state \cite{pratt_velocity-based_2006, xiong_3-d_2022}: the horizontal COM position with respect to the stance pivot $p$ and its velocity $v$  at the pre-impact event. Let $\x = [p, v]^T \in \R^2$ denotes the pre-impact horizontal COM state, the S2S dynamics of the robot is:
\begin{align}
    \x_{k+1} = \Poin_\x(x_k^-,\tau(t)).
\end{align}

\subsection{S2S Dynamics Approximation via Hybrid-LIP}
The continuous dynamics of the robot is highly nonlinear. Thus, we lack effective methods to obtain the analytical form of the robot S2S dynamics. A linear approximation to the robot S2S dynamics was proposed in \cite{xiong_3-d_2022} using a reduced-order model named Hybrid Linear Inverted Pendulum (H-LIP). With the vertical COM of the robot controlled to be approximately constant, the model discrepancy can be treated as bounded disturbances for a wide range of practically realizable walking behaviors. The S2S dynamics of the H-LIP model is as follows:
\begin{align}\label{eq::HLIPdynamics}
    \x_{k+1}^H = A^H \x_k^H + B^H u_k^H,
\end{align}
where $\x^H = [p^H, v^H]^T \in \R^2$ and $u^H \in \mathbb{R}$ represent the pre-impact horizontal COM state and the step size of the H-LIP. The dynamics matrix and actuation matrix of the H-LIP are denoted by $A^H \in \R^{2 \times 2}$ and $B^H \in \R^{2}$, respectively. See \cite{xiong_3-d_2022} for detailed derivation. The second coordinate $v$ can be replaced by the angular momentum about the contact pivot \cite{gong_angular_2021}, while the linear structure of the S2S dynamics remains the same. Using the H-LIP as an approximation, the robot S2S dynamics can be written as:
\begin{align}\label{eq::HLIPdynamicsApproximation}
    \x_{k+1} = A^H \x_k + B^H u_k + w_m, 
\end{align}
where $w_m = \Poin_\x(x_k^-,\tau(t))-A^H \x_k-B^H u_k$, and $u_k$ is the $k$-th step size of the robot. The term $w_m$ describes the model discrepancy, i.e., the integrated dynamics difference between the robot and the H-LIP over a step. Given the realizable set of walking behavior, $w_m \in \mathbf{W}_m$ where $\mathbf{W}_m$ is a bounded set \cite{xiong_3-d_2022}. Applying the H-LIP based stepping controller 
\begin{align}\label{eq::HLIPcontroller}
    u_k = u^H + K (\x_k - \x^H)
\end{align}
yields the error dynamics: $\mathbf{e}_{k+1} = (A^H +B^H K)\mathbf{e}_{k} + w_m$, where $\mathbf{e} := \x - \x^H$ is the error state and $K$ is the controller gain. Any selections of $K$ that result in stable $A^H+B^H K$ can drive $\mathbf{e}$ to converge to a positive invariant set $\mathbf{E}$, i.e. if $\e_k \in \mathbf{E}$, $\e_{k+1} \in \mathbf{E}$. In \cite{xiong_3-d_2022}, walking behaviors are realized on the robot by applying Eq. \eqref{eq::HLIPcontroller} using deadbeat controllers and linear quadratic regulators to design the feedback gain $K$.
This stepping controller has realized directional velocity tracking \cite{xiong_3-d_2022}, global position control \cite{xiong2021global} and walking on rough terrain \cite{xiong_slip_2021}.

\begin{figure*}[t]
  \centering\includegraphics[width=0.95\textwidth]{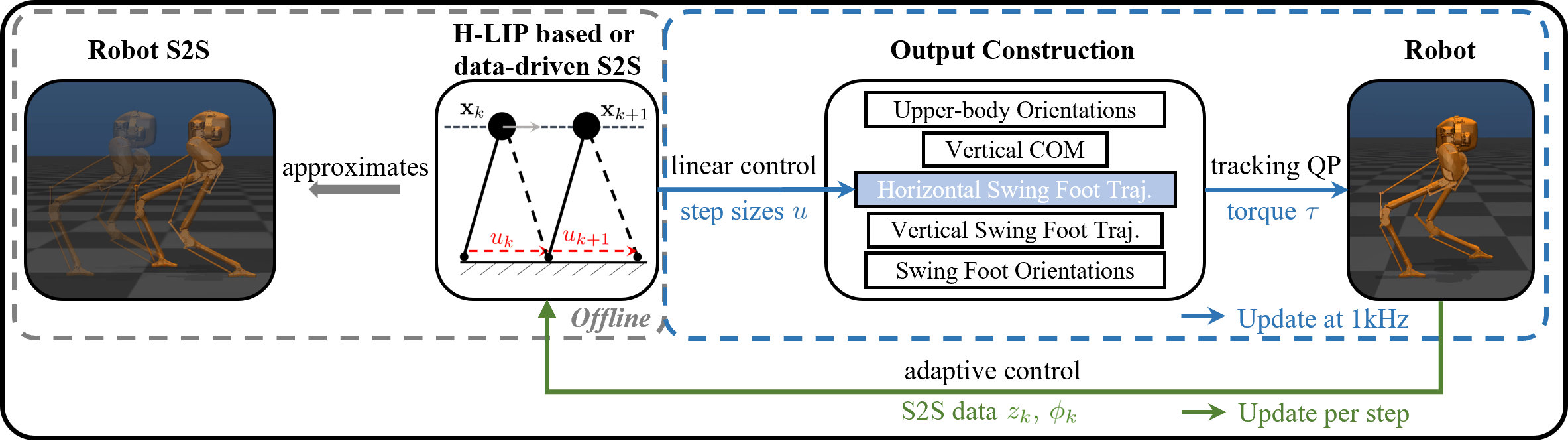}
  \vspace{-1mm}
  \caption{An overview of the proposed data-driven adaptive framework, where dashed boxes denote previous work \cite{xiong_3-d_2022, xiong_robust_2021}.}\label{fig::frameworkOVerview}
  \vspace{-5mm}
\end{figure*}

\subsection{Data-driven S2S Dynamics Approximation}\label{sec::prelim_data_driven}
The H-LIP based S2S approximation provides a baseline for realizing versatile bipedal walking, the data of which can be used to learn more accurate and robot-specific S2S dynamics. In \cite{xiong_robust_2021}, a large set of data, collected from walking robots with the original H-LIP controller, was used offline to fit a linear model of the following form:
\begin{align}\label{eq::LearnLinearModel}
    \x_{k} = A \x_{k-1} + B u_{k-1} + C.
\end{align}
To achieve desired 3D bipedal walking, the periodic behaviors of walking can be characterized into Period-1 (P1) and Period-2 (P2) orbits \cite{xiong_3-d_2022}, where 1 and 2 indicate the number of steps to complete a walking period.

\noindent \textbf{Period-1 Orbit:} The desired step size of a P1 orbit, denoted by $u^*$, is determined by the desired walking velocity $v^d$ and the step duration $T$: $u^* = v^d T$. The corresponding desired periodic pre-impact state for realizing $u^*$ is calculated by setting $\x_{k+1} = \x_k$ in Eq. \eqref{eq::LearnLinearModel}:
\begin{align}\label{eq::P1xstar}
    x^* = (I - A)^{-1} (B u^* + C),
\end{align}
where $I$ is the identity matrix. 

\noindent \textbf{Period-2 Orbit:} Unlike P1 orbits, the P2 orbit that realizes a desired velocity $v^d$ is not unique. Let the subscript $_{\text{L/}\text{R}}$ denote the left or right stance leg. The step sizes satisfy $u_\text{L}^* + u^*_\text{R} = 2 v^d T$. Selecting one step size determines the orbit. Solving $\x_{k+2} = \x_k$ yields the periodic pre-impact states:
\begin{align}\label{eq::P2xstar}
    \x_\text{L/R}^* = (I-A_\text{R/L} A_\text{L/R})^{-1} &( A_\text{R/L} B_\text{L/R} u_\text{L/R}^* + B_\text{R/L} u^*_\text{R/L} \nonumber \\
    &+ A_\text{R/L} C_\text{L/R} + C_\text{R/L} ) ,
\end{align}
using $\x_{k+1,\text{L/R}} = A_\text{R/L} \x_{k,\text{R/L}} + B_\text{R/L}u_{k,\text{R/L}} + C_\text{R/L}$. Different from the setting in \cite{xiong_robust_2021}, we are not assuming the left and right stance dynamics are identical in this work considering potential disturbances.

\textbf{Remark:} The state-feedback stepping controller in Eq. \eqref{eq::HLIPcontroller} can be directly applied for stepping stabilization using the data-driven S2S dynamics. One only needs to replace the nominal $u^H$ and $x^H$ of the H-LIP by the $x^*$ and $u^*$ of the data-driven S2S model: $x^*$ is calculated in Eq. \eqref{eq::P1xstar} for P1 orbits and in Eq. \eqref{eq::P2xstar} for P2 orbits.

\subsection{Output Construction and Stabilization}

The S2S-based stepping controller only provides discrete inputs (i.e. the step sizes) to the system. To realize the target step sizes, continuous desired horizontal trajectories of the swing foot are constructed using B\'ezier polynomials. To facilitate the linear S2S dynamics approximation, the desired vertical COM height $z_\text{com}^d$ is a fixed constant. The vertical desired trajectory of the swing foot is constructed based on the desired step duration $T$ to realize the desired lift-off and strike-down behaviors of walking. The rest of the desired trajectories include the orientations of the upper body and the swing foot, which can be set to constants. Finally, the desired walking trajectories are tracked using optimization-based controllers such as control Lyapunov function based quadratic programs (CLF-QP) \cite{ames_rapidly_2014} and task-space control based quadratic programs (TSC-QP) \cite{bouyarmane_quadratic_2019}.

\textbf{Remark:} Fig. \ref{fig::frameworkOVerview} overviews the previous framework in the dashed box. As shown in the solid box, the focus of this paper is to use adaptive control to learn the data-driven S2S model to improve the performance of the stepping controller. Thus, the output construction and the low-level tracking are kept intact as in \cite{xiong_3-d_2022}.

%% file: methodsnew.tex
\section{Online Adaptive S2S Dynamics Approximation based Control} \label{sec::adaptive}
To motivate the use of adaptive control, we first rewrite the S2S dynamics of the robot in Eq. \eqref{eq::LearnLinearModel} as 
\begin{align}
    \underbrace{\x_k}_{z_k} = \underbrace{\begin{bmatrix}
        A & B & C
    \end{bmatrix} }_{\Theta^{*T}} 
    \underbrace{\begin{bmatrix}
       \x_\km &
        u_\km &
        1
    \end{bmatrix}^T,}_{\phi_k}
\end{align}
where $z_k$, $\phi_k$ are available measurements, and $\Theta^{*}$ represents the unknown true model parameters. Adaptive control aims to iteratively solve for estimates of the true parameters, $\hTheta_k$,  using measurements as shown in Fig. \ref{fig::frameworkOVerview}. This section provides the general adaptive control framework and  our problem formulations, which include a state tracking formulation as in the motivating example and an output tracking formulation to account for tracking errors from low-level QP. 


\subsection{Adaptive Control}
As in the example, we use the classical linear static parametric model 
 \cite{ioannou_adaptive_2006} to describe the identification problem:
\begin{align}\label{eq::adaptive_thetastar}
    z_k = \Theta^{*T} \phi_k ,
\end{align}
where $z_k \in \R^m$ and $\phi_k \in \R^N$ are known signals, and $\Theta^{*} \in \R^{N \times m}$ represents the unknown true model parameters. The estimator is given by $\hat{z}_k = \hat{\Theta}_\km^{T} \phi_k$. The update law $\hTheta_k = g(\hTheta_\km,z_k,\phi_k)$ for the unknown parameters is determined by the adaptive control algorithm of the choice. In this work, we use the following update law called projection algorithm:
\begin{align}\label{eq::PA}
    \hTheta_k = \hTheta_{k-1} + \Gamma \phi_k (\phi_k^T \phi_k)^{-1} (z_k-\hTheta_{k-1}^T\phi_k)^T, \tag{PA}
\end{align}
where $\Gamma \in \R^{N \times N} \succ 0$ is a tunable gain. When $\Gamma = I_{N \times N}$, \eqref{eq::PA} is same as solving the following optimization problem:
\begin{align*}
   \underset{\hTheta_k} {\text{min}}  & \quad \frac{1}{2} \text{Tr}((\hTheta_k - \hTheta_\km)^T(\hTheta_k - \hTheta_\km) )  \\
\text{s.t.}  
 & \quad    z_k = \hTheta_k^T \phi_k,
\end{align*}
where $\text{Tr}()$ represents the matrix trace. Equivalently, the algorithm can be considered as iteratively solving an unconstrained optimization w.r.t. $\hTheta_k$ using the steepest gradient descent with the gain $\Gamma$ on the following cost function:
\begin{align*}
   J =  \frac{1}{2} \frac{\| z_k -\hTheta_k^T \phi_k \|}{ \phi_k^T \phi_k }  ,
\end{align*}
where the gradient is $\nabla J = \phi_k (\phi_k^T \phi_k)^{-1} z_k-\hTheta_{k-1}^T\phi_k^T$.

\vspace{0.2cm}
\noindent \textbf{Remark: } Many available adaptive control algorithms, such as the least-squared algorithm, solve the problem under cost functions with different formulations. 
Also, note that one could replace the vector $\phi_k$ with 
$\Phi_k = \begin{bmatrix}
    \phi_{k-q+1}& ... & \phi_{k-1} &\phi_k
\end{bmatrix} \in \R^{N \times q}.$
The use of $\Phi_k$ considers a horizon of data for optimization, rather than based on instantaneous cost as in \eqref{eq::PA}. In practice, we observed similar performance for the projection algorithm (basic and orthogonalized) and the least-square algorithm (weighted and unweighted). The proposed framework is not sensitive to the choice of algorithm and the projection algorithm presented is sufficient for different scenarios with minimal gain tuning.

\subsection{Data-driven S2S Formulation} \label{sec::formulation}

\subsubsection{State Tracking Formulation}\label{sec::basicformulation}
Consider the system dynamics in Eq. \eqref{eq::LearnLinearModel}, the estimator is given by
\begin{align}\label{eq::estimator}
    \hat{\x}_k &= \hat{A} \x_\km + \hat{B} u_\km + \hat{C}.
\end{align} 
The standard parameter identification form is:
\begin{align*}
z_k &= \x_k, \\
    \hTheta_k &= [\hat{A} \quad \hat{B} \quad \hat{C}]^T,\\
    \phi_k &= \begin{bmatrix} \x_{k-1}^T , u_{k-1}  , 1 \end{bmatrix}^T,
\end{align*}
where $\hTheta_k$ is initialized with the H-LIP model in Eq. \eqref{eq::HLIPdynamics} as: 
\begin{align}
    \hTheta_0 &= [A^H \quad B^H \quad 0_{2 \times 1}]^T.
\end{align}
Under the certainty equivalence principle \cite{ioannou_adaptive_2006}, we use the current estimate of the system model $\hTheta_k$ to synthesize the stabilizing controller:
\begin{align}\label{eq::Acontroller}
    u_k = u^* + K_k (\x_k - \hat{\x}^*_k),
\end{align}
where $u^*$, $\hat{\x}^*_k$, $K_k$ is calculated using the procedure provided in Sec. \ref{sec::prelim_data_driven} using the current estimates of $\hat{A}$, $\hat{B}$, $\hat{C}$. This controller stabilizes the data-driven reduced-order model to estimated fixed points.  
Note that for P2 orbits, the estimator and controller are dependent on the stance leg:
\begin{align*}
z_k &= \x_{k, \text{L/R}}, \\
    \hTheta_k &= [\hat{A}_\text{R/L} \quad \hat{B}_\text{R/L} \quad \hat{C}_\text{R/L}]^T,\\
    \phi_k &= \begin{bmatrix} \x_{k-1,\text{R/L}}^T , u_{k-1,\text{R/L}}  , 1 \end{bmatrix}^T,\\
    u_{k,\text{L/R}} &= u^*_\text{L/R} + K_{k,\text{L/R}} (\x_{k,\text{L/R}} - \hat{\x}^*_{k,\text{L/R}}),
\end{align*}
where the subscript of $K_{k,\text{L/R}}$ indicates the controller gain is solved w.r.t. the dynamics approximation of $\hat{A}_\text{L/R} , \hat{B}_\text{L/R} , \hat{C}_\text{L/R}$. 

\subsubsection{Output Tracking Formulation}\label{sec::outputformulation}
The previous formulations have only been focused on foot placement planning. In practice, the commanded step sizes are tracked closely but never exactly on the robot due to imperfect modeling and potential external disturbances. Thus, we present another formulation that accounts for the tracking error of the low-level QP. Let $y := u_\text{actual}$ be the actual step size that is realized by the robot, and $u := u_\text{cmd}$ is the planned step size. We assume a linear input-output relationship and use the following estimator:
\begin{align}\label{eq::output_linearform}
    \hat{y}_k = \hat{D} \x_k + \hat{E} u_k + \hat{F}.
\end{align}
With this, the adaptive estimator is provided by:
\begin{align*}
z_k &= [\x_k^T \text{, } y_{k-1}^T]^T, \\
    \hTheta_k &= \begin{bmatrix}
        \hat{A} & \hat{B} & \hat{C}\\\hat{D} & \hat{E} & \hat{F}
    \end{bmatrix}^T,\\
    \phi_k &= \begin{bmatrix} \x_{k-1}^T , u_{k-1}  , 1 \end{bmatrix}^T,
\end{align*}
where $\hTheta_k $ is initialized as 
\begin{align}
    \hTheta_0 &= \begin{bmatrix}A^H & B^H & 0_{2 \times 1} \\ 0_{2 \times 1 } & I_{1 \times 1 }& 0_{1 \times 1 }\end{bmatrix}^T.
\end{align}
The goal is to find $u_k$ such that $y_k$, the actual step size, converges to a reference $r$ while keeping $\x$ bounded. For our case, $r = u^*$. Thus, we could use the following controller:
\begin{align}\label{eq::AoutputTacking_controller}
    u_k = K_k \x_k + k_f r + b_f,
\end{align}
where $k_f$ and $b_f$ are calculated by substituting Eq. \eqref{eq::AoutputTacking_controller} into Eq. \eqref{eq::estimator} to find equilibrium $\x_e$. Then we solve $y_e = r$. If $r \neq 0$, $b_f = 0$, and if $r =0$, $k_f = 0$; otherwise,
\begin{align*}
    b_f &=  
        (M\hat{B}+\hat{E})^{-1} (r -  \hat{F} -M \hat{C} )  \quad\quad\  r = 0 \\
    k_f &= 
        r^{-1}(M\hat{B}+\hat{E})^{-1}(r - \hat{F}  - M\hat{C}) \quad  r \neq 0 
\end{align*}
where $M = (\hat{D}+\hat{E} \hat{K} )(I-\hat{A}-\hat{B} \hat{K})^{-1}$.



%% file: results.tex


\section{Results}\label{sec::results}

We evaluate the proposed approach on the robot Cassie \cite{cassie} built by Agility Robotics in our \texttt{C++} implementation in the open-sourced simulator \cite{cassie_mujocosim} with Mujoco physics engine \cite{todorov_mujoco_2012}. The procedure of the control implementation is summarized in Algorithm \ref{alg::framework}, and a flowchart is provided in Fig. \ref{fig::frameworkOVerview}. At every step, the desired step size is updated from the commanded velocity. Initialized from the H-LIP model, adaptive control updates the S2S model and solves the controller gain at every step of walking. During walking, continuous trajectories that realize the desired foot placement are then synthesized and tracked. Without losing generality, we select P1 orbit in the sagittal plane and P2 orbit in the lateral plane for the 3D orbit composition \cite{xiong_3-d_2022}. The efficacy and robustness of our walking method is evaluated in various scenarios listed below. In all cases except for scenarios in the velocity tracking section, we utilize a consistent $0.3$-second stepping cycle and $0.75$-meter stepping height for COM, which are parameters used for previous hardware experiments \cite{xiong_3-d_2022}. In addition, the same adaptive gain $\Gamma = 0.2 I$ is used across all cases. Visualization of the simulated walking can be seen in the supplementary video \cite{video-supp}. 

\begin{algorithm}[t]\caption{Adaptive S2S based Bipedal Walking}
\label{alg::framework}
 \begin{algorithmic}[1]
 \renewcommand{\algorithmicrequire}{\textbf{Initialization:}}
 \renewcommand{\algorithmicensure}{\textbf{Customization:}}
 \REQUIRE \textit{Gait Parameters}: Desired $z^d_\text{com}$, Walking Period $T$, B\'ezier polynomials. \textit{Adaptive Gain}: $\Gamma$.\\
\WHILE {Simulation}
\IF {Just Impacted}
\STATE Update $\hTheta_k$ using $\phi_k$ using \eqref{eq::PA}
\STATE Solve for the controller gain using $\hTheta_k$ 
\STATE Update $u^*$ from desired velocity and $x^*$ $\leftarrow$ Eq. \eqref{eq::P1xstar} and \eqref{eq::P2xstar} with current estimation of dynamics $\hTheta_k$
\ELSE 
    \FOR{every 1kHz}
\STATE Step Size Controller $\leftarrow$ Eq. \eqref{eq::Acontroller} or \eqref{eq::AoutputTacking_controller}\\
\STATE Continuous output synthesis \& low-level control 
\ENDFOR
\ENDIF \\

\ENDWHILE
 \end{algorithmic}
 \end{algorithm}

\subsection{Performance Improvement - Velocity Tracking}

\begin{figure}[t]
 \vspace{-0.3cm}
    \centering
    \includegraphics[width = \datafigurewidth]{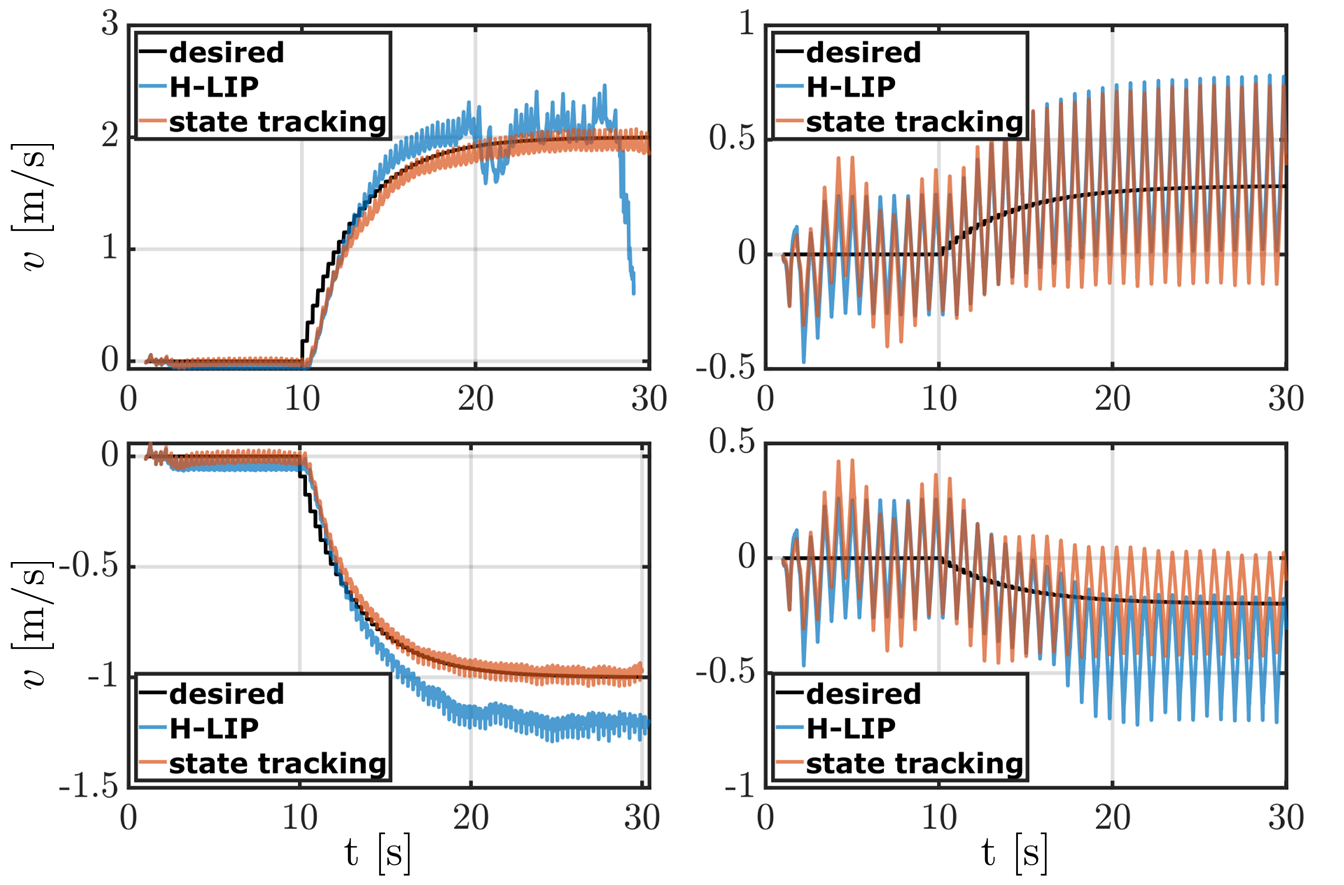}
     \vspace{-0.6cm}
    \caption{Comparison of velocity tracking performance of the baseline H-LIP controller vs. data-driven adaptive model-based state tracking controller: (left) sagittal velocity tracking with $T=0.3$s and $z_\text{com}=0.75$m; (right) lateral velocity tracking with $T=0.4$s, $z_\text{com}=0.65$m. }
    \vspace{-.6cm}\label{fig:vel_tracking_comparision}
\end{figure}
We first compare the velocity tracking performance between the H-LIP stabilization and the adaptive stabilization. One thing to note is that the performance of the H-LIP stabilization controller is heavily influenced by parameters such as the desired vertical COM height $z_\text{com}^d$ and the step duration $T$. Roughly speaking, a lower desired $z^d_\text{com}$ makes the walking more sensitive to the effect of leg swinging, and a longer walking duration increases the discrepancy between the integrated dynamics  $w_m$ of the robot and the H-LIP over the step; a larger dynamics difference $w_m$ produces larger state tracking errors under the original H-LIP based stepping controller. On the contrary, the adaptive control based framework stabilizes the robot to the desired walking speed regardless of the parameters used. As shown in Fig. \ref{fig:vel_tracking_comparision}, the velocity tracking under the H-LIP stepping varies under different gait parameters and commanded velocities; the proposed data-driven adaptive control noticeably improves the velocity tracking performance for both P1 and P2 orbits.


\subsection{Heavy Loads}\label{subsec::loads}

\begin{figure}[t]
    \centering
    \includegraphics[width = \datafigurewidth]{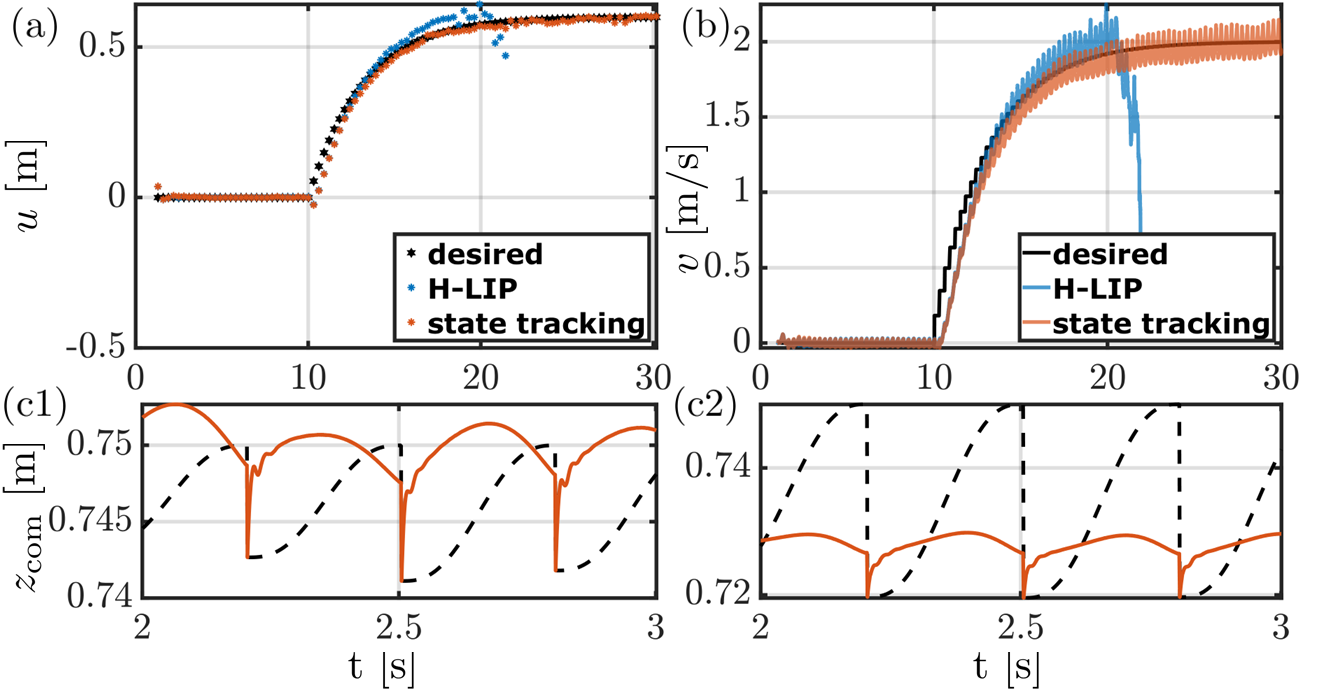}
    \vspace{-0.6cm}
    \caption{S2S based walking under unknown 10kg load: (a) commanded foot placement and (b) actual COM velocity. Vertical COM tracking $z_\text{com}$: (c1) without the external load, and (c2) with the 10kg load. The black dashed line is the desired output, and the red line is the actual output realized by the low-level controller.  }
    \vspace{-6mm}
    \label{fig::load}
\end{figure}

Next, we assess the adaptive S2S controller with additional weights on the robot. A box weighing 10kg, about $30 \%$ of the robot's own weight, is added to the pelvis of the robot. This load changes the COM kinematics and the whole-body dynamics of the robot. However, we assume none of the control synthesis procedures is aware of the load or tries to estimate the load. Instead, we expect the adaptive S2S dynamics to capture the influence of the load on the step level. As mentioned earlier, all remaining results have the same gait parameters: $z_\text{com}^d = 0.75$m and $T=0.3$s. As shown in Fig. \ref{fig::load}, the added mass has a negative influence on the vertical COM tracking. Nevertheless, our adaptive control still stabilized the system to the desired walking speed. We also included foot placement in Fig. \ref{fig::load}, which is an equivalent but less oscillatory metric for velocity tracking, as the average walking speed is simply total foot placement over 1 second. The nominal H-LIP based stepping fails to stabilize high-speed walking since the model inaccuracy exceeds its range of capability for realizing the bounded tracking error.  

\subsection{Modified Mass-Inertia Distributions}
\begin{figure}[t]
    \centering
    \includegraphics[width = \datafigurewidth]{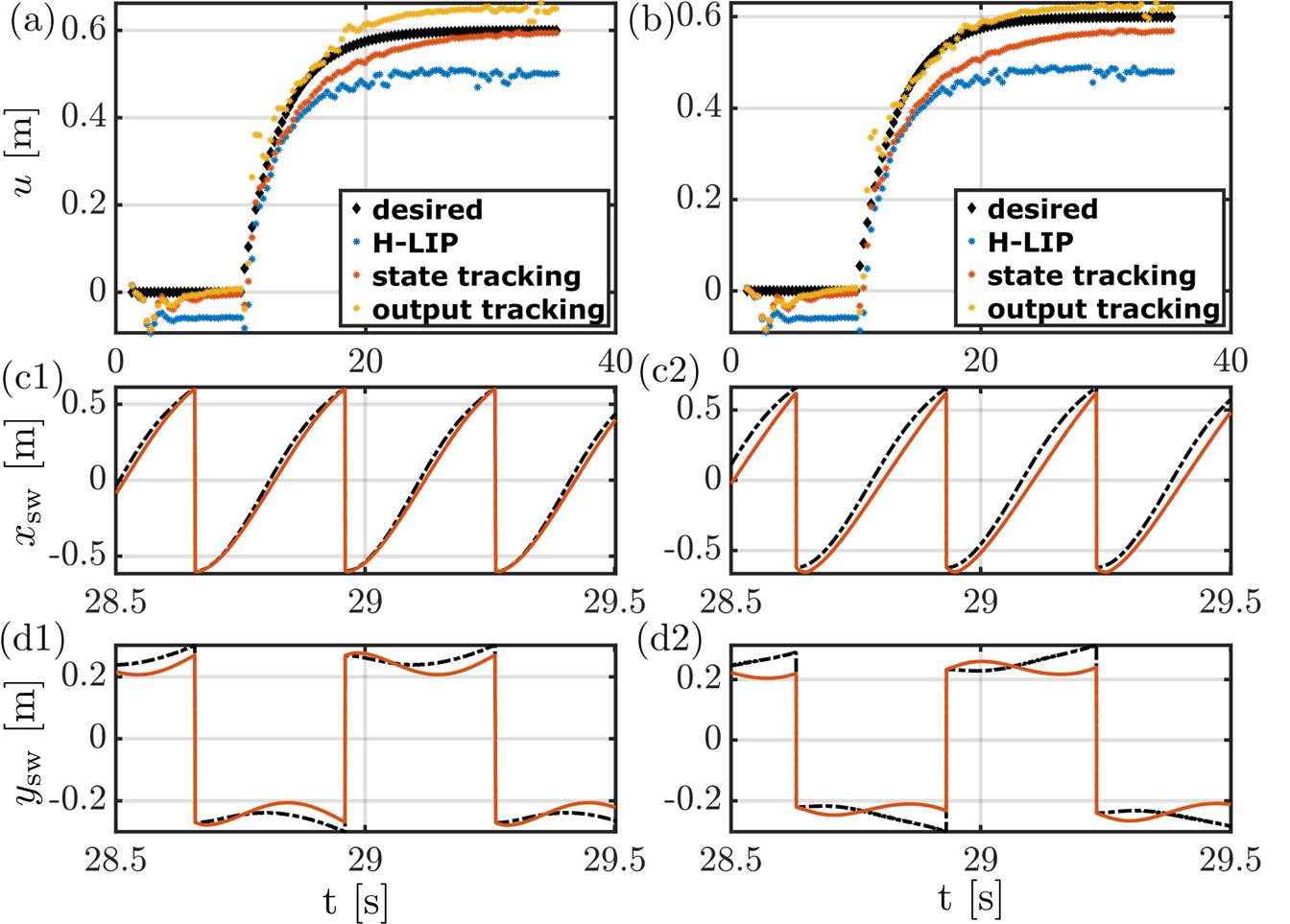}
    \vspace{-6mm}
    \caption{Comparison of two adaptive formulations for realizing walking under modified robot leg mass and inertia: commanded (a) and actual (b) sagittal step size. Horizontal swing foot trajectories $x_\text{sw}, y_\text{sw}$: (c1 \& d1) with nominal model, (c2 \& d2) with heavy leg model.}
    \vspace{-.3cm}\label{fig::formulation_comparision_leg_inertia}
\end{figure}
Besides the load on the top of the pelvis, we also test the controllers with increased leg mass and inertia (2.5 times these of the robot). The total mass of the robot is changed from 33.3kg to 45.2kg, which produces errors in COM kinematics and increases the tracking error on the horizontal swing foot trajectories as shown in Fig. \ref{fig::formulation_comparision_leg_inertia}. Similar to the previous cases, we expect the adaptive S2S dynamics to capture the dynamics difference without changing any gait construction or low-level controller. We use this scenario to compare the two different formulations: state tracking vs. output tracking. As shown in Fig. \ref{fig::formulation_comparision_leg_inertia}, the commanded step sizes in the state tracking formulation converge to the desired step sizes, but the actual step sizes still have a non-trivial steady-state error due to the low-level tracking errors coming from the inaccurate mass and inertia model of the robot. As we expect from the synthesis, the output tracking formulation directly reduces the tracking error of the actual foot placement and therefore provides better velocity tracking performance.

\subsection{State Estimation Error}
\begin{figure}[t]
    \centering
    \includegraphics[width = \datafigurewidth]{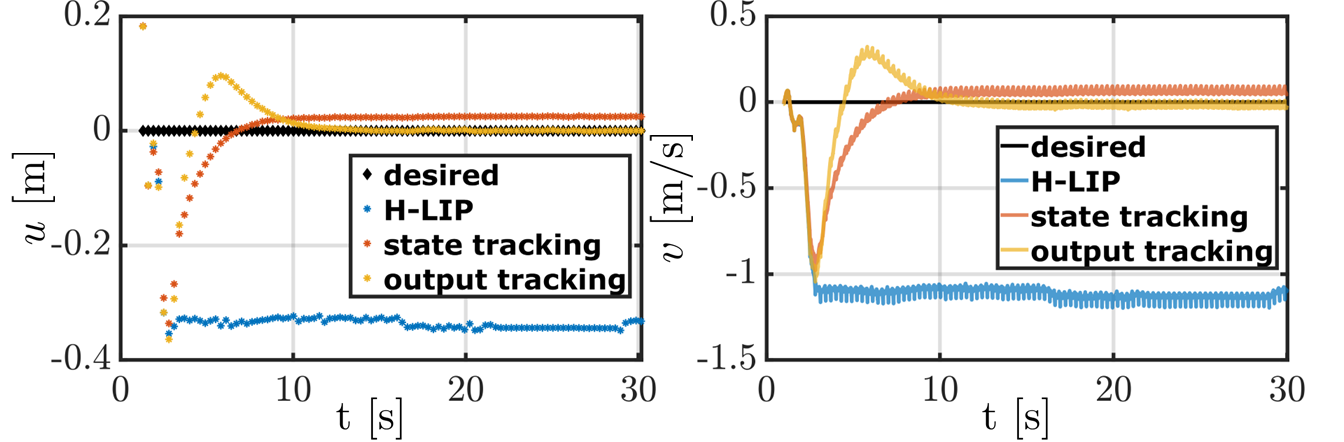}
    \vspace{-6mm}
    \caption{Walking velocity tracking under biased state estimation: actual foot placement (left) and actual COM velocity (right) .}
    \vspace{-.5cm}
    \label{fig::estimation}
\end{figure}
\begin{figure}[t]
    \centering
    \includegraphics[width = 0.95 \datafigurewidth]{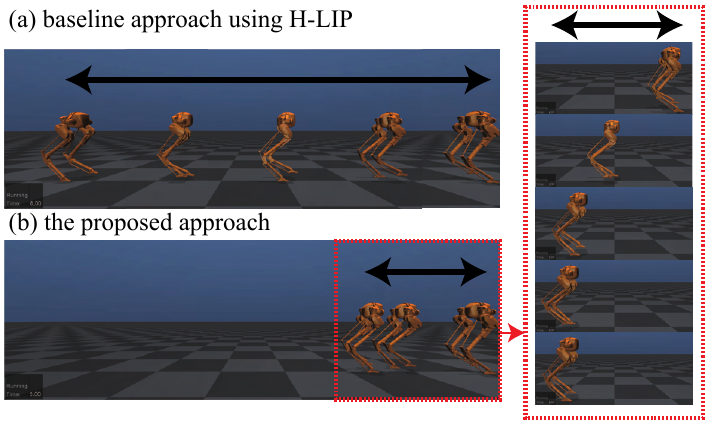}
    \vspace{-2mm}
    \caption{Snapshots of simulations for velocity tracking under biased state estimation: (a) baseline approach using H-LIP, (b) the proposed state-tracking approach. The snapshots shown are within 6s. }
    \vspace{-.3cm}
    \label{fig::estimation_snaps}
\end{figure}

The linear velocities of the floating base, i.e., the pelvis, of the robot cannot be directly measured in practice. Thus, for the experimental realization of walking controllers, sensor fusion for state estimation \cite{hartley2018hybrid, gao2022invariant} has to be implemented on the hardware to estimate the pelvis velocity, which may produce certain biased estimation errors. We virtually add a constant offset of $0.4$m/s to the state estimation of the forward horizontal velocity of the COM. Fig. \ref{fig::estimation} shows the step sizes and the velocity in the sagittal plane under different stepping controllers, and Fig. \ref{fig::estimation_snaps} provides snapshots of the simulation. For H-LIP based one, the robot converged to a periodic walking behavior but with a much larger velocity error than the added velocity estimation bias $0.4$m/s. The new fixed point can be found similarly to the orbit characterization. Let $b \in \R^2 $ be the estimation bias, the H-LIP controller is then $u^H = K(\x + b - \x^H) + u^H$. The equilibrium point $\x_e$ is solved by setting $\x_\kp = \x_k$ for Eq. \eqref{eq::LearnLinearModel}. The resulting fixed point under the estimation bias $b$ is given by:
\begin{align*}
    \x_e = (I - A - BK)^{-1}(BKb - BK \x^H + Bu^H + C).
\end{align*}
This agrees with the simulation results with $b=[0 \quad 0.4]^T$. In the given scenario, the absolute velocity estimation error of $0.4$m/s exaggerates to around $1.1$m/s error for steady-state velocity tracking. With the adaptive controllers, despite the initial oscillations due to the bias, both the step size and the walking velocity converge to the desired ones closely, and the output tracking formulation on the S2S dynamics has even better velocity tracking performance.

\subsection{External Drag Forces}
\begin{figure}[t]
    \centering
    \includegraphics[width = \datafigurewidth]{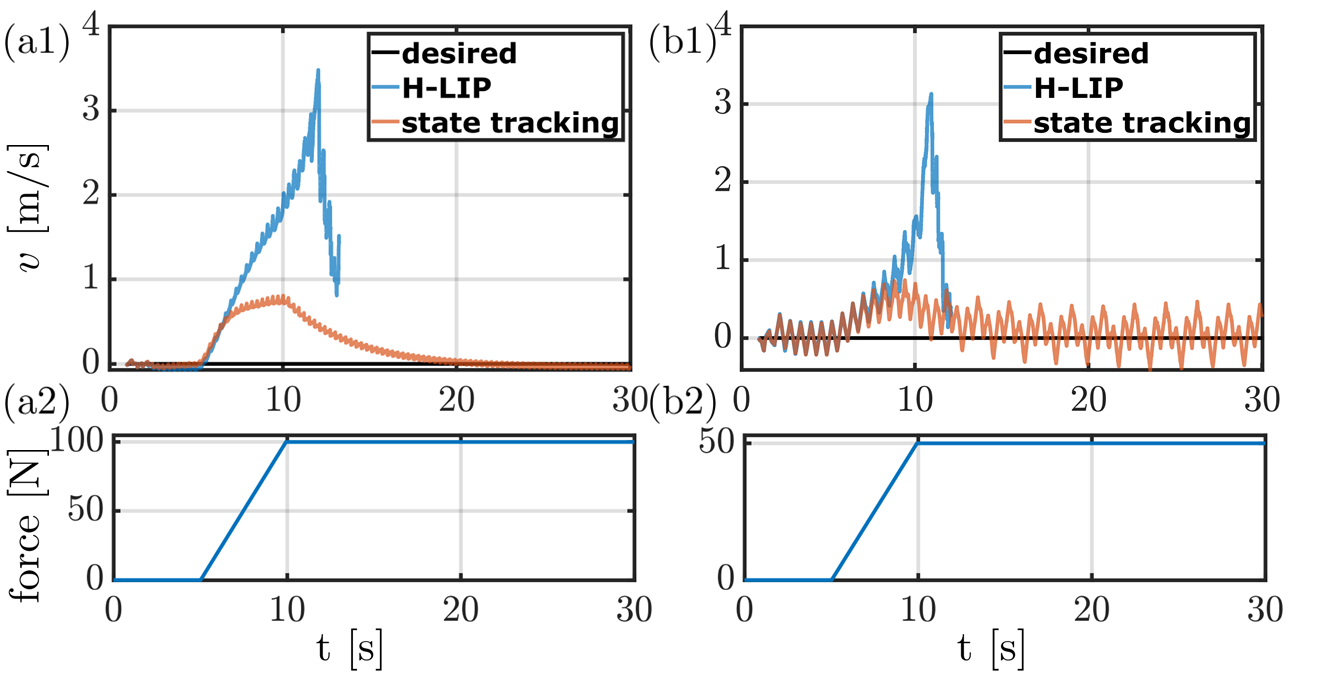}
    \vspace{-6mm}
    \caption{Velocity tracking performance and the corresponding force profiles under (a) sagittal  and (b) lateral  drag forces.}
    \vspace{-5mm}
    \label{fig::disturbforce}
\end{figure}

\begin{figure}[t]
    \centering
    \includegraphics[width = \datafigurewidth]{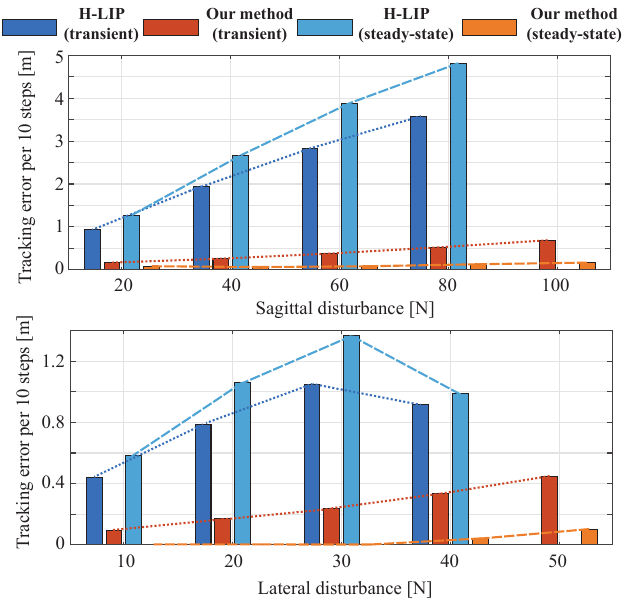}
    \caption{Tracking error analysis for different disturbance magnitude. Note that the baseline H-LIP approach presents higher steady-state error than the transient error as disturbance is gradually increased to steady-state disturbance during the transient phase.}
    \vspace{-.6cm}
    \label{fig::disturbforce_histogram}
\end{figure}
Moreover, we show the adaptive controller is able to handle persistent external drag forces in horizontal directions. A horizontal linear drag force is applied to the pelvis of the robot while the robot is commanded to step in place. The applied force is set to increase from 0 to a constant value in 5s. One method of assessing the resilience of bipedal robots is to incrementally amplify the magnitude of disturbances until the robots lose their balance. This approach has been widely used to demonstrate the efficacy of controllers such as \cite{lee_online_2022}, same procedure is used to evaluate the resilience of our approach. As shown in Fig. \ref{fig::disturbforce}, we apply up to $100$N in the sagittal plane and $50$N in the lateral plane. With the nominal H-LIP stepping, the walking speed of the robot continuously increases as the drag force increases, and the robot eventually falls over. With the stabilization performed with the adaptive data-driven controller, the robot finds its new equilibrium and converges to stable periodic stepping-in-place motion. 

Fig. \ref{fig::disturbforce_histogram} compares the transient and steady-state error with different magnitudes of applied disturbance forces. To ensure a fair evaluation, the assessment is divided into two distinct horizons: the transient phase (comprising the first $80$ steps) and the steady-state phase (comprising $80$-$100$ steps). We then calculate the tracking errors, which is the travel distance from the start normalized by the time horizon of $3$ seconds (per $10$ steps). As we can see from Fig. \ref{fig::disturbforce_histogram}, both the transient and steady-state error for baseline approaches are much larger than the state-tracking adaptive formulation. The tracking error for the H-LIP approach increases almost linearly as the disturbance increases, except that lateral $40$N disturbance leads to a tilted yaw angle so part of the error is in the sagittal plane.  The results of the analyses demonstrate that our method effectively enhances the robustness of bipedal walking, particularly under larger perturbing forces.

\subsection{Slopes} \label{subsec::slopes}

\begin{figure}[t]
    \centering
    \includegraphics[width = \datafigurewidth]{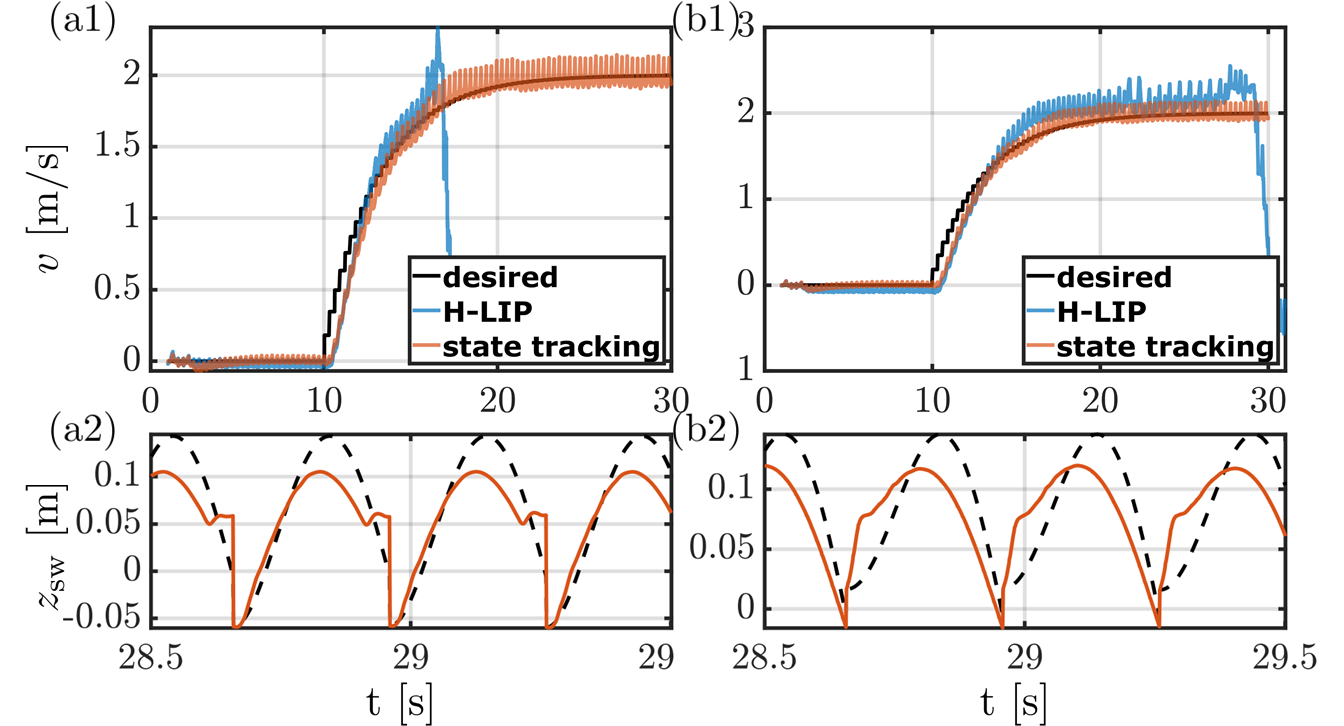}
    \vspace{-0.6cm}
    \caption{Walking velocity tracking on up- (a1) and down-slopes (b1), and vertical swing foot trajectory $z_\text{sw}$ on up- (a2) and down-slopes (b2).}
    \vspace{-.6cm}
    \label{fig::slope}
\end{figure}

Finally, it is noteworthy that the adaptive S2S controller can handle walking on unknown sloped terrains. We test the performance of our method on slopes of $\pm 0.1$ rad, where the desired foot angle is not zero but rather set to be parallel to the stance foot at each step for the low-level tracking. Since the vertical swing foot trajectory is constructed based on time, the slope directly affects the actual impact time as shown in the swing foot vertical trajectories in Fig. \ref{fig::slope}, where early impact occurs for an upward slope and late impact occurs for a downward slope. Furthermore, we observe that the error of impact time is positively correlated to the commanded step size. This results in the failure of the nominal H-LIP controller for dynamic walking at high speeds, whereas our data-driven formulation adapts to the change and successfully tracks the desired velocity. 

\subsection{Discussion}\label{sec::discussion}
The proposed method extends the previous offline data-driven approach \cite{xiong_robust_2021} with comparable results observed for the general velocity tracking task without additional disturbances. However, the offline method cannot handle any disturbances that are not presented during the data collection period. By incorporating online features, the proposed adaptive S2S controller demonstrates superior performance in all different disturbance scenarios listed in Sec. \ref{subsec::loads} to \ref{subsec::slopes}. As shown in the figures, when the velocity reference reaches its steady state, the resulting walking motions converge to the desired reference within 10 seconds for all cases. Faster convergence can be achieved by increasing the adaptive gain but leads to a higher risk of inducing unstable transient behavior. As the adaptive control is applied at the S2S level, there is no need to identify the type of disturbance or estimate its magnitude. In our approach, the impact of different disturbances is all embedded into the high-level step-to-step dynamics, leading to effortless and low-dimensional data-efficient implementation. 

The method is primarily independent of the low-level tracking controller, except for the output tracking formulation that takes into account the actual foot placement, which is a result of the low-level controller. In practice, we found out the output tracking formulation is much more sensitive to the choice of adaptive gain and the speed of changing reference velocity. An inappropriate selection of these parameters results in undesirable oscillations during the transient phase. On the contrary, the state tracking formulation is very robust to any of the parameter changes. We hypothesize that the linear equation on tracking error used in Eq. \eqref{eq::output_linearform} is not a good description, as the sensitivity disappears when we take the commanded foot placement as the output, i.e., $D = [0 \quad0] $, $E=1$, $F=0$. Therefore, future research should explore alternative methods to integrate the tracking error of the low-level controller into our framework. We plan to report experimental results shortly.

For the sake of space, we only evaluated the proposed methodology with persistent disturbances. Impulse-type disturbances that do not affect the steady-state walking behavior are omitted. This type of disturbance can be rejected under the current framework using the proposed state feedback-based foot placement control as in \eqref{eq::Acontroller}. The allowable range for impulse disturbance is confined by the robot's kinematic and dynamic feasibility. 

 
\section{Conclusion}\label{sec::conclusion}

In conclusion, we have proposed an online framework that utilizes a data-driven step-to-step dynamics model to synthesize bipedal walking behaviors that adapt to changes in the environment and robot dynamics. 
This layered method addresses the complex control problems of bipedal walking in an adaptive fashion. Our approach is shown to improve the robustness of dynamic walking and the tracking performance of walking velocity through simulations under various disturbance scenarios.
